\documentclass[letterpaper]{article} % DO NOT CHANGE THIS
\usepackage[preprint]{neurips_arxiv}
\usepackage[T1]{fontenc}
\usepackage[latin9]{inputenc}
\usepackage{caption}
\usepackage{array}
\usepackage{multirow}
\usepackage{amsmath}
\usepackage{amsthm}
\usepackage{amssymb}
\usepackage{stackrel}
\usepackage{graphicx}
\usepackage{algorithm,algpseudocode}
\algnewcommand\algorithmicforeach{\textbf{for each}}
\algdef{S}[FOR]{ForEach}[1]{\algorithmicforeach\ #1\ \algorithmicdo}

\begin{document}

\title{Improving Long Handwritten Text Line Recognition with Convolutional Multi-way Associative Memory}

\author{\textbf{Duc Nguyen}\\
Cinnamon AI Lab, Vietnam\\
john@cinnamon.is
\and
\textbf{Nhan Tran}\\
Cinnamon AI Lab, Vietnam\\
lucas@cinnamon.is
\and
\textbf{Hung Le}\\
Deakin University, Australia\\
lethai@deakin.edu.au
}

\maketitle
\begin{abstract}
Convolutional Recurrent Neural Networks (CRNNs) excel at scene text
recognition. Unfortunately, they are likely to suffer from vanishing/exploding
gradient problems when processing long text images, which are commonly
found in scanned documents. This poses a major challenge to goal of
completely solving Optical Character Recognition (OCR) problem.
Inspired by recently proposed memory-augmented neural networks (MANNs)
for long-term sequential modeling, we present a new architecture dubbed Convolutional Multi-way
Associative Memory (CMAM) to tackle the limitation of current CRNNs.
By leveraging recent memory accessing mechanisms in MANNs, our architecture
demonstrates superior performance against other CRNN counterparts
in three real-world long text OCR datasets. 
\end{abstract}

\section{Introduction}

Since first introduced, CRNN for Handwritten Text Recognition (HTR) has been constantly breaking state-of-the-art results \cite{doetsch2014fast,pham2014dropout,voigtlaender2016handwriting}, and being deployed in industrial applications \cite{bluche2017gated,borisyuk2018rosetta}. The gist of CRNN involves a convolutional feature extractor that encodes visual details into latent vectors, followed by a recurrent sequence decoder that turns the latent vectors into human-understandable characters. The whole architecture is trained end-to-end via Connectionist Temporal Classification (CTC) loss function \cite{graves2006connectionist} or attention mechanism \cite{bahdanau2015neural,hori2017advances,kim2017joint}.

\begin{figure}
\begin{centering}
\includegraphics[width=0.7\textwidth]{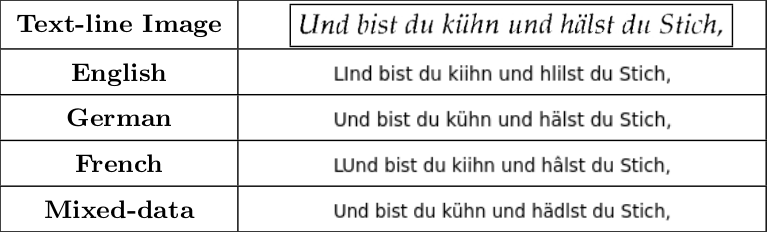}
\caption{Models trained on English, German, French, Mix-of-3 give prediction on a German text line (taken from \cite{ul2013can})\label{fig:Language-Model}}
\end{centering}
\end{figure}

Inside CRNN, the role of the sequence decoder (often implemented as Long Short-Term Memory \citep{hochreiter1997long}) has been reported to serve as a language model \cite{sabir2017implicit}. In \cite{sabir2017implicit}, the authors observe that a learned OCR model attains higher accuracy for meaningful text line than for random text line. Experimental results from \cite{ul2013can} also support this claim, in that an OCR model performs worse when tested on languages other than the language it is trained on (see Fig. \ref{fig:Language-Model}). These results intuitively make sense because the knowledge of surrounding characters can provide clues to ascertain correct prediction. We hypothesize that this effect is even more pronouncing for HTR, where handwriting style variations and real world conditions can render characters visually confusing and unrecognizable, making predictions of such characters only feasible by referring to the surrounding context. As a result, enhancing the sequence decoder would improve CRNN performance.

Despite the promising empirical results, the LSTM as well as other RNN-based models is  incapable of remembering long context due to vanishing/exploding gradient problems \cite{bengio1994learning,le2018learning,pascanu2013difficulty}. Current HTR datasets, which contain only scene text or short segments of texts are not challenging enough to expose the weakness. However, when approaching industrial data, which often involve transcription of the whole line of documents written in complicated languages, RNN-based decoders may fail to achieve good results.

Recently, a new class of neural network architecture, called Memory-Augmented Neural Network (MANN), has demonstrated potentials to replace RNN-based methods in sequential modeling. In essence, MANNs are recurrent controller networks equipped with external memory modules, in which the controllers can interact with external memory unit via attention mechanisms \cite{graves2014neural}. Compared to LSTM, instead of storing  hidden states in a single vector of memory cells, a MANN can store and retrieve its hidden states in multiple memory slots, making it more robust against exploding/vanishing gradient problem. MANNs have been experimented to perform superior to LSTM in language modeling tasks \cite{gulcehre2017memory,kumar2016ask} and thus beneficial to HTR where language modelling  supports recognition.  However, for this problem, to the best of our knowledge, there is no work currently employing MANN.

In this work, we adapt recent memory-augmented neural networks by integrating an external memory module into a convolutional neural network. The CNN layers read the input image, encoding it to a sequence of visual features. At each timestep of the sequence, two controllers will be used to store the features into the memory, access the memory in multiple refinement steps and generate an output. The output will be passed to a  CTC layer to produce the CTC loss and the final predicted character. In summary, our main contributions are:

\begin{enumerate}
    \item We introduce a memory-augmented recurrent convolutional architecture for OCR, called Convolutional Multi-way Associative Memory (CMAM),
    \item We demonstrate the new architecture's performance on 3 handwritten datasets: English IAM, Chinese SCUT-EPT and a private Japanese dataset.
\end{enumerate}

\section{Related Work}

\subsection{Convolutional Recurrent  Neural Networks}

Attempts to improve CRNN have mostly been focused on the convolutional encoder. Starting with the vanilla implementation \cite{shi2017end}, convolution layers have since been incorporated with gating mechanism \cite{bluche2017gated}, recurrent connections \cite{wang2017gated,lee2016recursive}, residual connections \cite{borisyuk2018rosetta,gao2017reading,zhan2017handwritten}, and used alongside with dropout \cite{pham2014dropout,puigcerver2017multidimensional}, or Maxout \cite{he2016reading}. Multi-Dimensional LSTM (MDLSTM) \cite{graves2007multi}, which is often used in the feature encoder, can also serve as the sequence decoder \cite{graves2009offline,pham2014dropout,voigtlaender2016handwriting}. On the other hand, improvement on the sequence decoder has received little attention. \cite{sun2017gmu} proposes GMU, inspired by the architectures of both LSTM and GRU \citep{chung2014empirical}, and achieves good results on both online English and Chinese handwriting recognition tasks. \cite{doetsch2014fast} controls the shape of gating functions in LSTM with learnable scales. Recently, there are some attempts that replace CTC with Seq2Seq decoder, both without \cite{sahu2015sequence} and with attention mechanism \cite{bluche2017scan} support for decoding. Hybrid models between CTC and attention decoder \cite{hori2017advances,kim2017joint} are also proposed and gain big improvement in Chinese and Japanese speech recognition.

\subsection{Memory-augmented Neural Networks}
Memory-augmented neural networks (MANN) have emerged as a new promising research topic in deep learning. In MANN, the interaction between the controller and the memory is differentiable, allowing it to be trained end-to-end with other components in the neural network \cite{graves2014neural,weston2014memory}. Compared to LSTM, it has been shown to generalize better on sequences longer than those seen during training \cite{graves2014neural,graves2016hybrid}. This improvement comes at the expense of more computational cost. However, recent advancements in memory addressing mechanisms allow MANN to perform much more efficiently \cite{le2018learning}. For practical applications, MANNs have been applied to many problems such as meta learning \cite{santoro2016meta}, healthcare \cite{10.1007/978-3-319-93040-4_22,Le:2018:DMN:3219819.3219981}, dialog system \cite{le2018variational}, process analytic \citep{khan2018memory} and extensively in question answering \cite{miller2016key} and language modeling \cite{gulcehre2017memory}. Our work (CMAM) is one of the first attempts to utilize MANN for HTR tasks.

\section{Proposed System}

\subsection{Visual Feature Extraction with Convolutional Neural Networks\label{cnn_sec}}

\begin{figure*}
\begin{centering}
\includegraphics[width=\textwidth]{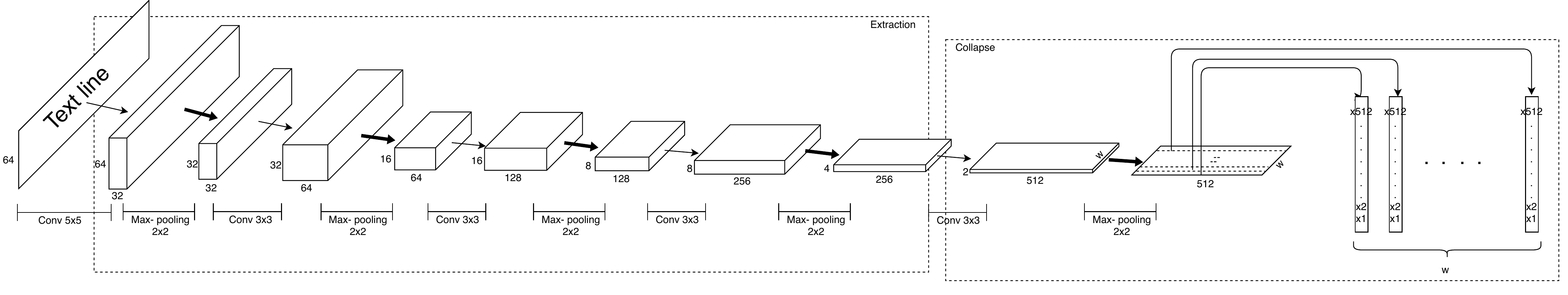}
\par\end{centering}
\caption{Convolutional layers\label{fig:Convolutional-layers}}
\end{figure*}

In our CMAM, the component of convolutional layers is constructed by
stacking the convolutional and max-pooling layers as in a standard CNN
model. Such component is used to extract a sequential feature representation
from an input image. Before being fed into the network, all the images
need to be scaled to the same height. Then a sequence of feature vectors
is extracted from the feature maps produced by the component of convolutional
layers, which is the input for the memory module. Specifically, each
feature vector of a feature sequence is generated from left to right
on the feature maps column by column. This means the $i$-th feature vector
is the concatenation of the $i$-th columns of all the maps (see Fig.
\ref{fig:Convolutional-layers}). Each of the feature vector is then fed
to a fully-connected layer to produce the final input $\mathbf{x}_{t}$
for the memory module.

\subsection{Sequential Learning with Multi-way Associative Memory}

In this section, we propose a new memory-augmented neural network, 
namely Multi-way Associative Memory (MAM) that is designed for HTR
tasks. 

\subsubsection{Memory-augmented neural network  overview}

A memory-augmented neural network (MANN) consists of an LSTM controller,
which frequently accesses and modifies an external memory $M\in\ensuremath{\mathbb{R}}^{N\times D}$,
where $N$ and $D$ are the number of memory slots and the dimension
of each slot, respectively. At each time step, the controller receives
a concatenation of the input signal $x_{t}$
and the previous read values $r_{t-1}$ from the memory to update
its hidden state and output as follows,

\begin{equation}
h_{t},o_{t}=LSTM\left(\left[x_{t},r_{t-1}\right],h_{t-1}\right)\label{eq:h_t}
\end{equation}

The controller output $o_{t}$ then is used to compute the memory
interface $\xi_{t}$ and the short-term output $y_{t}^{s}$ by applying
linear transformation,

\begin{equation}
\xi_{t}=W_{\xi}o_{t}
\end{equation}
\begin{equation}
y_{t}^{s}=W_{s}o_{t}
\end{equation}
where $W_{\xi}$,$W_{s}$ are trainable weight parameters. The interface
$\xi_{t}$ in our specific model is a set of vectors $\left\{ k_{t}^{rj},\beta_{t}^{rj},f^{rj},k_{t}^{w},\beta_{t}^{w},v_{t},e_{t},g_{t}^{alc},g_{t}^{w}\right\} $
responsible for controlling the memory access, which includes memory reading
 and writing procedures (see \ref{subsec:Memory-reading} and
\ref{subsec:Memory-writing}, respectively). After the memory is accessed,
the read values for current step $r_{t}$ are computed and combine
with the short-term output $y_{t}^{s}$, producing the final output
of the memory module as the following,

\begin{equation}
y_{t}=W_{y}\left[y_{t}^{s},r_{t}\right]\label{eq:y_t}
\end{equation}
where $W_{y}$ is trainable weight parameter. With the integration
of read values, the final output $y_{t}$ contains not only short-term
information from the controller, but also long-term context from the
memory. This design was first proposed in \cite{graves2014neural}
and has become a standard generic memory architecture for sequential
modeling \cite{graves2014neural,graves2016hybrid,le2018variational, 10.1007/978-3-319-93040-4_22,Le:2018:DMN:3219819.3219981}. 

\subsubsection{Multi-way Associative Memory}

We leverage the standard memory-augmented architecture by marrying
bidirectional control with multi-hop memory accesses. In particular,
we use two controllers: forward and backward controllers (both implemented
as LSTM). The forward controller reads the inputs in forward order
(from timestep $1$-th to $T$-th) together with the read contents
from the memory. It captures both short-term information from the
past and long-term knowledge from the memory. On the other hand, the
backward controller reads the inputs in backward order (from timestep
$T$-th to $1$-th) and thus only captures short-term information
from the future. The backward controller maybe useful when the short-term
future timesteps give local contribution to recognize the character at current timestep.
However, it is unlikely that long-term future timesteps can have causal
impact on current temporary output, which explains why the backward
controller does not need to read content from the memory. 

Moreover, our architecture supports multi-step computations to refine
the outputs from the controllers ($r_{t}$ and $y_{t}^{s}$) before
producing the final output for that timestep. Let us denote $L$ as the
number of refinement steps. At $l$-th refinement step, the short-term
output of previous refinement $y_{t,l-1}^{s}$ will be used as the
input for the controllers. In particular, the controllers will compute their hidden
states and temporary outputs for this refinement as follows,

\begin{equation}
h_{t,l}^{f},o_{t,l}^{f}=LSTM^f\left(\left[y_{t,l-1}^{s},r_{t-1,l}\right],h_{t-1,l}^{f}\right),t=1,...,T
\end{equation}

\begin{equation}
h_{t,l}^{b},o_{t,l}^{b}=LSTM^b\left(y_{t,l-1}^{s},h_{t+1}^{b}\right),t=T,...,1
\end{equation}
where $y_{t,-1}^{s}=x_{t}$. At timestep $t$-th, the forward controller
updates its hidden state $h_{t,l}^{f}$ and compute temporary output
$o_{t,l}^{f}$. The temporary output is stored in a buffer, waiting
for the backward controller\textquoteright s temporary output $o_{t,l}^{b}$.
The two buffered outputs are used to compute the memory interface
vector $\xi_{t,l}$ and the short-term output $y_{t,l}^{s}$ as the
following,

\begin{equation}
\xi_{t,l}=W_{\xi}\left[o_{t,l}^{f},o_{t,l}^{b}\right]
\end{equation}
\begin{equation}
y_{t,l}^{s}=W_{s}\left[o_{t,l}^{f},o_{t,l}^{b}\right]
\end{equation}

\begin{figure}
\begin{centering}
\includegraphics[width=0.8\textwidth]{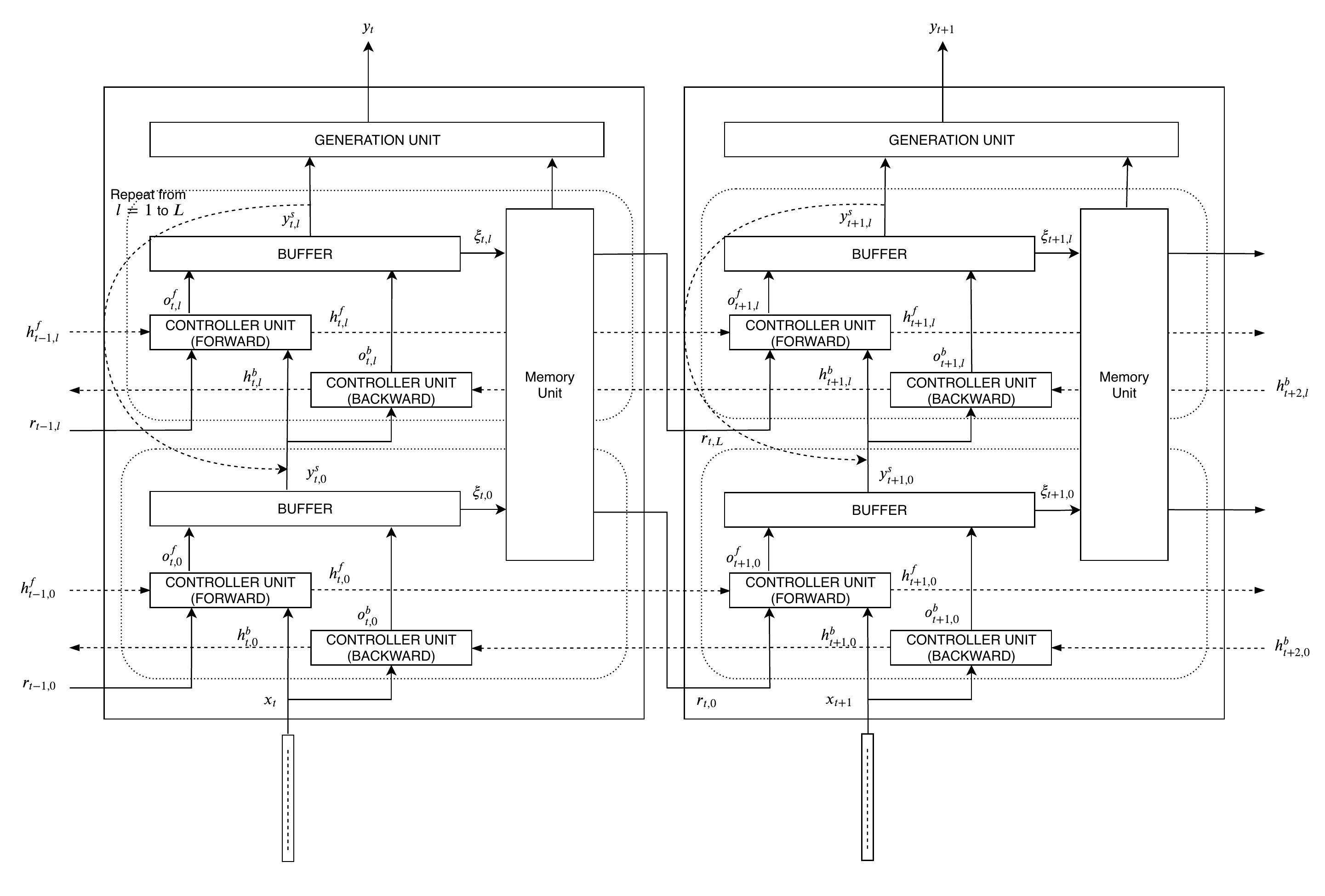}
\par\end{centering}
\caption{Multi-way Associative Memory Module. At each timestep, the two controllers collect information from the recent past and future, together with long-term values from the memory to make predictions. The memory is accessed in both horizontal and vertical ways (hence the name multi-way). \label{fig:Associative-Memory-Module}}
\end{figure}

The refinement process simulates human multi-step reasoning. We often
refer to our memory many times before making final decision. At each
step of refinement, the controllers re-access the memory to get information
representing the current stage of thinking. The representation is
richer than the raw representation stored in the memory before refinement.
For example, without refinement, at timestep $t$-th, the forward
controller can only read values $r_{t,0}$ containing information
from the past. However, from the first refinement step ($l\geq1$), the
memory is already filled with the information of the whole sequence,
and thus a new refined read at timestep $t$-th can contain (very
far if necessary) future information $r_{t,l}$. Fig. \ref{fig:Associative-Memory-Module}
illustrates the flow of operations in the architecture. 

After refinement process, the final output $y_{t}$ is computed by
the generation unit as follows,

\begin{equation}
y_{t}=W_{y}\left[y_{t,L}^{s},r_{t,L}\right]\label{eq:y_t-1}
\end{equation}

We summarize the operation of MAM in Alg. \ref{alg:Multi-layer-bidirectional-memory}.
The algorithm can be considered as a generalization of the bidirectional
control memory design proposed in \cite{W18-2606}. 

\subsubsection{Memory reading\label{subsec:Memory-reading}}

In our MAM, memory reading is inspired by content-based attention mechanism
presented in \cite{graves2014neural,graves2016hybrid}. Let $R$ be
the number of reading heads, the read keys $k_{t}^{rj}\in\ensuremath{\mathbb{R}}^{D},j=\overline{1,R}$
will be used for locating the $j$-th read-out. The addressing mechanism
is mostly based on cosine similarity measure,

\begin{equation}
D\left(M_{t}(i),k_{t}^{rj}\right)=\frac{k_{t}^{rj}\cdot M_{t}(i)}{||k_{t}^{rj}||\cdot||M_{t}(i)||}
\end{equation}
which is used to produce a content-based read-weight $w_{t}^{rj}\in\ensuremath{\mathbb{R}}^{N}$
whose elements are computed according to  softmax function over memory's locations,

\begin{equation}
w_{t}^{rj}(i)=\text{softmax}\left(D\left(M_{t}(i),k_{t}^{rj}\right)\beta_{t}^{rj}\right)
\end{equation}
Here, $\beta_{t}^{rj}$ is the strength parameter. After the read
weights are determined. the $j$-th read value $r_{t}^{j}$ is retrieved
as the following,

\begin{equation}
r_{t}^{j}=\stackrel[j=1]{R}{\sum}w_{t}^{rj}(i)M_{t}(i)
\end{equation}
The final read-out is the concatenation of all read values $r_{t}=\left[r_{t}^{1},r_{t}^{2},...,r_{t}^{R}\right]$.

Different from \cite{graves2016hybrid}, we exclude temporal linkage
reading mechanism from our memory reading. Recent analyses in \cite{W18-2606}
reveal that this mechanism increases the computation time and physical
storage dramatically. By examining the memory usages, \cite{W18-2606}
also indicated that the temporal linkage reading are barely used in
realistic tasks and thus could be safely removed without hurting the
performance much. We follow this practice and only keep the content-based
reading mechanism. We realize that the key-value based retrieval resembles traditional associative memory system \citep{baird1993neural}. The mechanism is
critical for OCR tasks where reference to previous exposures of some
visual feature may consolidate the confidence of judging current ones.
When compared to recurrent networks in current OCR systems, which
only use single visual feature stored in the hidden state to make
prediction on the output, multiple visual features reference may provide
the model with richer information and thus give better predictions. 

\begin{algorithm}
\begin{algorithmic}[1]
\small
\Require{Given initial $r_{0,l}$, $h^f_{0,0}$, $h^b_{T+1,0}$, forward LSTM cell $LSTM^f$, backward LSTM cell $LSTM^b$, memory functions $read()$, $write()$ and input sequence $[x_1,x_2,...,x_T]$}
\State{Set $[y_{1,-1}^{s},y_{2,-1}^{s},...,y_{T,-1}^{s}]=[x_1,x_2,...,x_T]$}
\For{$l=0,L$}\Comment{$L$ is the number of refinement steps}
\For{$t=T,1$}\Comment{$T$ is the number of timesteps}
\State{$o^b_{t,l},h^b_{t,l}=LSTM^b([y_{t,l-1}^{s},h^b_{t+1,l})$}
\EndFor
\For{$t=1,T$}
\State{$o^f_{t,l},h^f_{t,l}=LSTM^f([y_{t,l-1}^{s},r_{t-1}],h^f_{t-1,l})$}
\State{$\xi_{t,l}=W_{\xi}\left[o_{t,l}^{f},o_{t,l}^{b}\right]$}
\State{$y_{t,l}^{s}=W_{s}\left[o_{t,l}^{f},o_{t,l}^{b}\right]$}
\State{$write(\xi_{t,l})$}
\State{$r_{t,l}=read(\xi_{t,l})$}
\EndFor
\EndFor
\end{algorithmic} 

\caption{Multi-way Memory Access Algorithm\label{alg:Multi-layer-bidirectional-memory}}
\end{algorithm}

\subsubsection{Memory writing\label{subsec:Memory-writing}}

To build writing mechanism for our memory, we apply three different
writing strategies. First. we make use of the dynamic memory allocation
in \cite{graves2016hybrid}. This strategy tends to write to least-used
memory slots. Let use the memory retention vector $\psi_{t}\in\left[0,1\right]^{N}$
to determine how much each memory location will not be freed as follows,

\begin{equation}
\psi_{t}=\stackrel[j=1]{R}{\prod}\left(1-f_{t}^{j}w_{t-1}^{rj}\right)
\end{equation}
where for each read head $j$, $f_{t}^{j}\in\left[0,1\right]$ denotes
the free gate emitted by the interface and $w_{t-1}^{rj}$ denotes
the read weighting vector from the previous timestep. The usage over
$N$ memory locations at current time-step $t$ is given by $u_{t}\in\left[0,1\right]^{N}$,
which is called memory usage vector,

\begin{equation}
u_{t}=\left(u_{t-1}+w_{t-1}^{w}-u_{t-1}\circ w_{t-1}^{w}\right)\circ\psi_{t}
\end{equation}
Then, the allocation vector is defined as the following,

\begin{equation}
a_{t}\left[\varPhi_{t}\left[k\right]\right]=\left(1-u_{t}\left[\varPhi_{t}\left[k\right]\right]\right)\stackrel[i=1]{k-1}{\prod}u_{t}\left[\varPhi_{t}\left[i\right]\right]
\end{equation}
in which, $\varPhi_{t}$ contains elements from $u_{t}$ in sorted
order.

The second strategy we propose is the last-read writing, in which
the location to be written is the previous read location. We define
the previous read locations by averaging the previous read weights,

\begin{equation}
l_{t}=\frac{1}{R}\stackrel[j=1]{R}{\sum}w_{t-1}^{rj}
\end{equation}
By writing to the previous read address, we assume that after read,
the content in that address is no longer important for future prediction.
This assumption makes sense in OCR setting where some visual features
only take part in recognizing one character. After the model refers
to these visual features to make predictions, it is safe to remove
them from the memory to save spaces for other important features. 

The third strategy is the common content-based writing, which is similar
to content-based reading. A content-based write weight is computed
as follows,

\begin{equation}
c_{t}^{w}(i)=\text{softmax}\left(D\left(M_{t}(i),k_{t}^{w}\right)\beta_{t}^{w}\right)
\end{equation}
where $k_{t}^{w}$ and $\beta_{t}^{w}$ are the key and strength parameters
for content-based writing, respectively.

To allow the model to select amongst strategies and have the ability
to refuse writing, a write mode indicator $g_{t}^{alc}\in\ensuremath{\mathbb{R}}^{3}$
and a write gate $g_{t}^{w}\in\ensuremath{\mathbb{R}}$ are used to
compute the final write weight as the following,

\begin{equation}
w_{t}^{w}=g_{t}^{w}\left[g_{t}^{alc}\left(0\right)a_{t}+g_{t}^{alc}\left(1\right)l_{t}+g_{t}^{alc}\left(2\right)c_{t}\right]
\end{equation}
where the write mode indicator $g_{t}^{alc}$ and the write gate $g_{t}^{w}$
are normalized using softmax and sigmoid functions, respectively.

Finally, the write-weight can be used together with the update value
$v_{t}\in\ensuremath{\mathbb{R}}^{D},$and erase value $e_{t}\in\left[0,1\right]^{D}$
to modify the memory content as follows,

\begin{equation}
M_{t}=M_{t-1}\circ\left(E-w_{t}^{w}e_{t}^{\top}\right)+w_{t}^{w}v_{t}^{\top}
\end{equation}
where $\circ$ is element-wise product.

\section{Experimental Evaluation}

\begin{table}
\begin{centering}
\begin{tabular}{ccccccc}
\hline 
\multirow{2}{*}{Dataset} & \multirow{2}{*}{Language} & \multirow{2}{*}{\#Writers} & \multicolumn{2}{c}{\#Lines} & Avg. characters & \multirow{2}{*}{\#Classes}\tabularnewline
\cline{4-5} 
 &  &  & Train+valid & Test & per line & \tabularnewline
\hline 
IAM & English & 400 & 7,458 & 2,915 & 45.72 & 80\tabularnewline
SCUT-EPT & Chinese & 2986 & 40,000 & 10,000 & 26.41 & 4,058\tabularnewline
Private & Japanese & N/A & 14,594 & 2,940 & 49.83 & 2,227\tabularnewline
\hline 
\end{tabular}
\par\end{centering}
$\\$
\caption{Characteristics of the datasets used in our experiments.\label{tab:Characteristics-of-the}}

\end{table}

\begin{figure}
\begin{centering}
\includegraphics[width=0.7\textwidth]{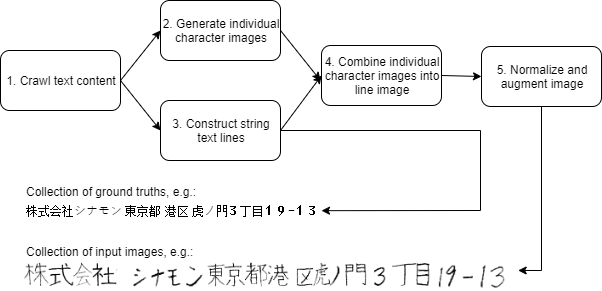}
\par\end{centering}
\caption{Synthetic Line Image Generation\label{fig:Synthetic-Line-Image-Generation}}
\end{figure}

\subsection{Experimental Settings}

\paragraph{Baselines.}
The main baseline used across experiments is the traditional CRNN with the same CNN architecture presented in \ref{cnn_sec}, coupled with the bidirectional 1D LSTM layers as proposed in \citep{puigcerver2017multidimensional}. We use DNC  \citep{graves2016hybrid} as the decoder for visual sequence to form another MANN baseline against our memory-based architecture. Finally, we also include other results reported from previous works for SCUT-EPT dataset
\paragraph{Datasets.}To validate our proposed model, we select IAM \citep{marti2002iam} and SCUT-EPT \citep{zhu2018scut}, which are two public datasets written in English and Chinese, respectively. We also collect a private dataset from our partner-a big corporation in Japan. This dataset is scanned documents written by Japanese scientists and thus, contains noises, special symbols, and more characters per line than the other datasets. The statistics of the three datasets are summarized in Table \ref{tab:Characteristics-of-the}.

Since our aim is to recognize long text line by end-to-end models, we do not segment the line into words. Rather, we train the models with the whole line and report Character Error Rate (CER), Correct Rate (CR) and Accuracy Rate (AR) \citep{yin2013icdar}. To fit with Chinese and Japanese datasets where there is no white-space between words, we exclude white-space from the vocabulary set and do not measure word error rate metrics.

\subsection{Synthetically generated handwritten line images\label{subsec:synthetic_data}}
This section describes the synthetic data generation process used for our experiments.
We execute 5 steps as illustrated in Fig. \ref{fig:Synthetic-Line-Image-Generation}.
Specifically, we start by crawling some text contents, typically from news site or from Wikipedia, and remove bad symbols and characters from the text corpus (1). Since the corpus might contain unknown characters to us, i.e. characters that we do not have any visual knowledge. We obtain these characters (a) by scouting them on the web and (b) by generating them from fonts and apply random variations to make it less print-like and more handwriting-like (2). The corpus also largely contains text of paragraph level, so we break it down into text lines, each of which contains on average 15 characters (3). For each text line, we combine random individual character images to make line images (4). Finally, we normalize the line images to ease the abrupt variations produced by different image characters and we augment the line images to increase style variations. After this process, we have a collection of line images with their corresponding ground truth labels.
\begin{table}
\begin{centering}
\begin{tabular}{ccc}
\hline 
\multirow{1}{*}{Model} & \multicolumn{1}{c}{Valid} & \multicolumn{1}{c}{Test}\tabularnewline
\hline 
CRNN (BiLSTM) & \multirow{1}{*}{13.19$\pm$0.24} & \multirow{1}{*}{13.72$\pm$0.16}\tabularnewline
DNC & 14.15$\pm$0.47 & 14.61$\pm$0.34\tabularnewline
\hline 
CMAM ($l=0$) & \multirow{1}{*}{12.99$\pm$0.40} & \multirow{1}{*}{13.39$\pm$0.15}\tabularnewline
CMAM ($l=1$) & \multirow{1}{*}{\textbf{10.72$\pm$0.23}} & \multirow{1}{*}{\textbf{11.12$\pm$0.20}}\tabularnewline
\hline 
\end{tabular}
\par\end{centering}
$\\$

\caption{Character Error Rate (CER) on IAM dataset\label{tab:iamfull}}
\end{table}

There are two purposes of using synthetic data. First, we generate data for tuning our implemented models (CRNN, DNC, CMAM). In particular, after training with 10,000 synthesized line images (both Latin and Japanese) on various range of hyper-parameters, we realize that the optimal memory size for MANN models is 16$\times$16  with 4 read heads. The LSTM controller's hidden sizes for DNC and CMAM are 256 and 196, respectively. For the CRNN baseline, the best configuration is 2-layer bidirectional LSTM of size 256. The optimal optimizer is RMSprob with a learning rate of $10^{-4}$. Second, since collecting and labeling line images are labor-intensive, we increase the number of training images with synthetic data for the Japanese recognition task. We generate 100,000 Japanese line images to pre-train our models before fine-tuning them on the private Japanese dataset.
\subsection{Latin Recognition Task}

In this task, we compare our model with CRNN and DNC. Two variants of our CMAM are tested: one uses no refinement ($l=0$) and the other uses one step of refinement ($l=1$), respectively. This experiment is a simple ablation study to determine good configuration of CMAM for other tasks. We run each model 5 times and calculate the mean and  standard deviation on CER metric. The final result is reported in Table \ref{tab:iamfull}. 

Compared to other baselines, DNC is the worst performer, which indicates a naive application of this generic model on OCR tasks seem inefficient. Both versions of CMAM outperform other baselines including the common CRNN architecture. Increasing the number of refinement steps helps CMAM outperform CRNN more than 2$\%$. The improvement is not big since the IAM's text line is short and clean.
It should be noted that our results are not comparable to that reported in other works \citep{puigcerver2017multidimensional} as we train on the whole line and discard white-space prediction. We also do not use any language model, pre-processing and post-processing techniques.

\subsection{Chinese Recognition Task}

\begin{table}
\begin{centering}
\begin{tabular}{ccc}
\hline 
\multirow{2}{*}{Model} & \multirow{2}{*}{CR} & \multirow{2}{*}{AR}\tabularnewline
 &  & \tabularnewline
\hline 
CRNN (LSTM) & 78.60 & \textbf{75.37}\tabularnewline
Attention & 69.83 & 64.78\tabularnewline
Casacade Attention & 54.09 & 48.98\tabularnewline
CRNN (BiLSTM, ours) & 81.47 & 74.33\tabularnewline
\hline 
CMAM ($l=1$) & \textbf{82.14} & 74.45\tabularnewline
\hline 
\end{tabular}
\par\end{centering}
$\\$

\caption{Correct Rate (CR) and Accuracy Rate (AR) on raw SCUT-EPT dataset.\label{tab:scut}}
\end{table}
In this experiment, we run CRNN and CMAM ($l=1$) on SCUT-EPT dataset. Table \ref{tab:scut} shows the CR and AR measurements of our implemented models and other baselines from \cite{zhu2018scut}. CRNNs are still  strong baselines for this task as they reach much higher accuracies than attentional models. Compared to CRNN \citep{zhu2018scut}, our proposal CMAM can demonstrate higher CR by more than 3$\%$, yet lower AR by 0.92$\%$. However, CMAM outperforms our CRNN implementation on the two metrics by 0.67$\%$ and 0.12$\%$, respectively.

\subsection{Japanese Recognition Task}

The OCR model is trained on 2 sources of datasets: synthetically generated handwritten line images (see Sec. \ref{subsec:synthetic_data}) and a private in-house collected Japanese dataset. The dataset contains more than 17,000 handwritten line images, which is divided into training and testing sets (see Table \ref{tab:Characteristics-of-the}). We split 2000 images from the training set to form the validation set. These images are obtained from scanned notebooks, whose lines are meticulously located and labeled by our dedicated QA team. It contains various Japanese character types (Katakana, Kanji, Hiragana, alphabet, number, special characters and symbols) to make our model work on general use cases.

We have conducted 2 experiments with the Japanese dataset. In the first experiment, we train the models directly with the real-world training set and report CER on the validation and testing set in Table \ref{tab:Results-on-Showadenko-1} (middle column). As seen from the results, CMAM outperforms CRNN significantly where the error rates reduce around 14$\%$ in both validation and testing sets. This demonstrates the advantage of using memory to capture distant visual features in case the line text is long and complicated. 

In the second experiment, we first train the models on 100,000 synthetic lines of images. After the models converge on the synthetic data, we continue the training on the real-world data. The result is listed in the rightmost column of Table
\ref{tab:Results-on-Showadenko-1}. With more training data, both models achieve better performance. More specifically, the improvement can be seen clearer in the case of CRNN, which implies pre-training may be more important for CRNN than CMAM. However, CMAM can still reach lower error rates than CRNN on the validation and test set (9$\%$ and 1$\%$, respectively). 

\begin{table}
\begin{centering}
\begin{tabular}{ccccc}
\hline 
\multirow{2}{*}{Model} & \multicolumn{2}{c}{Original data} & \multicolumn{2}{c}{Enriched data}\tabularnewline
 & valid & test & valid & test\tabularnewline
\hline 
CRNN (BiLSTM, ours) & 32.84  & 27.00 & 24.83 & 11.62\tabularnewline
\hline 
CMAM ($l=1$) & \textbf{17.55} & \textbf{12.99} & \textbf{15.66} & \textbf{10.71}\tabularnewline
\hline 
\end{tabular}
\par\end{centering}
$\\$

\caption{Character Error Rate (CER) on Japanese dataset\label{tab:Results-on-Showadenko-1}}
\end{table}

\section{Conclusion}

In this paper, we present a new architecture for handwritten text recognition that augments convolutional recurrent neural network with an external memory unit. The memory unit is inspired by recent memory-augmented neural networks, especially the Differentiable Neural Computer, with extensions for new writing strategy and multi-way memory access mechanisms. The whole architecture is proved efficient for handwritten text recognition through three experiments, in which our model demonstrates competitive or superior 
performance against other common baselines from the literature. 

\bibliography{uw}
\bibliographystyle{plain}

\end{document}